\documentclass[10pt, a4paper]{article}

\usepackage{lrec2026} 
\usepackage{fancyvrb}
\usepackage{multirow}
\usepackage{subcaption}
\definecolor{lightblue}{rgb}{0.8,0.9,1.0}  
\definecolor{forest}{HTML}{ABEB9E}  
\definecolor{brightgreen}{HTML}{5CBCE6}   
\definecolor{bordeaux}{HTML}{5CBCE6}  
\definecolor{lightorange}{HTML}{FFED80} 

\title{From Polyester Girlfriends to Blind Mice: Creating the First Pragmatics Understanding Benchmarks for Slovene}

\name{Mojca Brglez, Špela Vintar}

\address{Jožef Stefan Institute \\   Jamova 39, Ljubljana, Slovenia
\\ \{mojca.brglez, spela.vintar\}@ijs.si\\ \\
 Faculty of Arts, University of Ljubljana \\ Aškerčeva 2, Ljubljana, Slovenia}

\abstract{
Large language models are demonstrating increasing capabilities, excelling at benchmarks once considered very difficult. As their capabilities grow, there is a need for more challenging evaluations that go beyond surface-level linguistic competence. Namely, language competence involves not only syntax and semantics but also pragmatics, i.e., understanding situational meaning as shaped by context as well as linguistic and cultural norms. To contribute to this line of research, we introduce SloPragEval and SloPragMega, the first pragmatics understanding benchmarks for Slovene that contain altogether 405 multiple-choice questions. We discuss the difficulties of translation, describe the campaign to establish a human baseline, and report pilot evaluations with LLMs. Our results indicate that current models have greatly improved in understanding nuanced language but may still fail to infer implied speaker meaning in non-literal utterances, especially those that are culture-specific. We also observe a significant gap between proprietary and open-source models. Finally, we argue that benchmarks targeting nuanced language understanding and knowledge of the target culture must be designed with care, preferably constructed from native data, and validated with human responses.
 \\ \newline \Keywords{large language models, benchmarking, pragmatics, dataset creation} }

\begin{document}

\maketitleabstract

\section{Introduction}

Large language models are approaching human levels of performance in several tasks. Generative AI is marked by a discourse-like setting: typical use cases involve turn-taking between a user and an agent, transforming LLMs into conversational partners. It is thus important to assess the capabilities of language models in understanding its users, as mutual understanding has large consequences for successful communication and can potentially influence the performance on many other downstream tasks. 

To truly assess the level of understanding or linguistic competence in LLMs, more difficult and complex tasks are needed,  i.e., those that require more than just the grasp of surface linguistic structures. In humans, language competence goes beyond mastering the surface structure (syntax) and meaning (semantics); it also entails an understanding of how context, along with linguistic and cultural norms, contributes to the situational meaning (pragmatics). 
The latter is created from and influenced by context in the widest possible sense, including the speakers, listeners, cultural norms, social norms, individual experience, communicative setting, what is said, and also what is not said. Pragmatics is thus concerned with language that is non-literal, context-dependent, inferential, and/or not truth-conditional \citep{birner_pragmatics_12}. All of these levels of language may contribute to what is called "nuanced language", i.e., context-sensitive language that marks subtle distinctions in meaning, tone, or stance, often via pragmatic resources.


\begin{figure*}[ht!]%
 \centering%
 \begin{subfigure}{0.5\linewidth}%
 \centering%
 \includegraphics[width=0.8\linewidth]{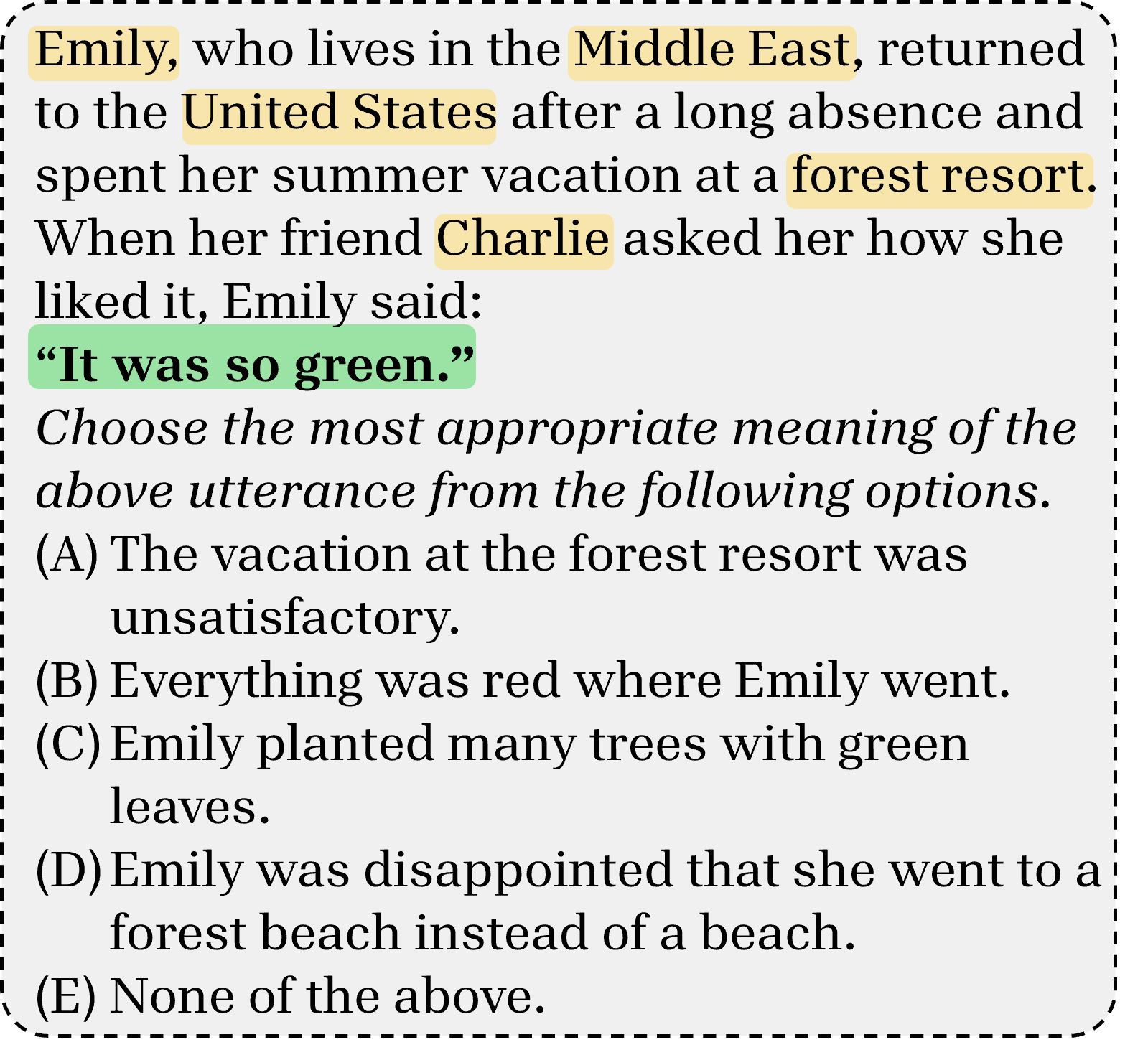}%
 \caption{}\label{fig:ex_1a}%
 \end{subfigure}\begin{subfigure}{0.5\linewidth}%
 \centering%
 \includegraphics[width=0.8\linewidth]{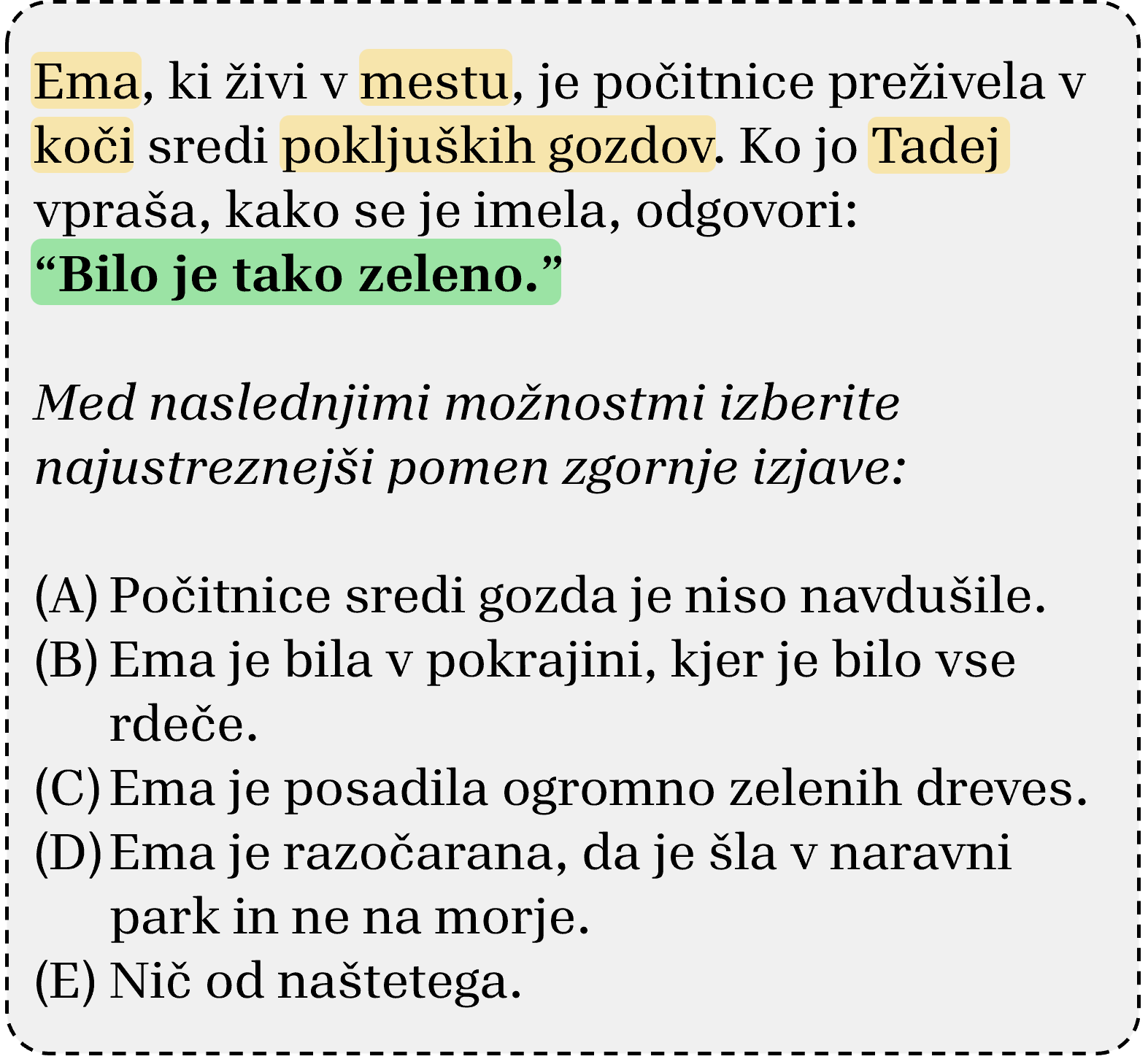}%
  \caption{}%
  \label{fig:ex_1b}%
 \end{subfigure}%
 \caption{Example of a Quantity-flouting utterance from MultiPragEval (a) and SloPragEval (b). Culturally specific terms in yellow. Utterance in green, bolded.}%
 \label{fig:example_1MPE}%
\end{figure*}

Researchers have recently begun targeted evaluations of pragmatic reasoning and nuanced language understanding \citep{park-etal-2024-multiprageval,sravanthi-etal-2024-pub,wu-etal-2024-rethinking}. Many studies have shown that LLMs still struggle to understand some nuanced language phenomena, such as irony or faux pas \citep{hu-etal-2023-fine,strachan2024}. Secondly, they face even greater difficulties when moving outside English \citep{park-etal-2024-multiprageval}, which is unsurprising given findings that LLMs are culturally biased towards the Western Anglo-Saxon space, in particular the US \citep[e.g.,]{yao_jue_performance_biases_24,zhou-etal-2025-mapo,alkhamissi-etal-2024-investigating}.

To evaluate the usefulness of those same LLMs for other, smaller languages, it is important to create benchmarks that accommodate both the linguistic and cultural context of the target language. Most of the current practice of creating non-English benchmark datasets relies on machine translation, sometimes with no post-editing involved. Specifically, upon examination of some existing machine-translated benchmarks, we argue that this approach produces culturally maladapted datasets unsuitable for non-literal language, leading to synthetic, unreliable evaluations.

In our work, we address the gap in evaluating the capabilities of LLMs in understanding various types of nuanced Slovene language. We translate and adapt previously used pragmatics benchmarks to create SloPragEval\footnote{Available on the SloBench benchmarking platform at \url{https://slobench.cjvt.si/leaderboard/view/15}} and SloPragMega\footnote{Available on the SloBench benchmarking platform at \url{https://slobench.cjvt.si/leaderboard/view/16}}. 
We discuss the problems of machine translation and the need for careful design of datasets involving rounds of revision and testing on natural speakers before using the data as a benchmark reference.


\section{Related Work}

\paragraph{Benchmarks for Nuanced Language}

With the growing use of large language models (LLMs) in conversational AI and other discourse-like scenarios, a series of probes and benchmarks have been introduced to test their communicative abilities. Pragmatic understanding requires a vast array of capabilities and knowledge, from mastering language, social and cultural knowledge, to mental-state reasoning. 
Thus, many datasets exist for the evaluation of specific linguistic abilities directly or indirectly related to pragmatics. In these, models are evaluated on a particular downstream task, such as the identification of metaphors (\citealp[e.g.,]{boisson-etal-2025-automatic}), understanding irony \citep[e.g.,]{WenTian+2025+259+283}, or natural language inference/entailment tasks \citep[e.g.,]{halat-atlamaz-2024-implicatr}. 
\citet{sileo2022pragmaticscenteredevaluationframeworknatural} present PragmEval, one of the first comprehensive benchmarks for English pragmatic understanding, integrating 11 datasets. 
Similarly, more recent benchmarking practices combine a variety of tasks to assess social, emotional, and pragmatic reasoning, which frequently connect with long-standing psychodiagnostic or psychometric tests. For example, \citet{choi-etal-2023-llms} create 58 tasks to evaluate what they refer to as "social knowledge", testing humor, sarcasm, offensiveness, sentiment, emotion, and trustworthiness. On the other hand, in addressing Theory of Mind capabilities in LLMs, \citet{jones-etal-2024-comparing-humans} find that LLMs display considerable sensitivity to mental states and match human performance in several tasks. However, they also find systematic errors in other tasks, especially those requiring pragmatic reasoning on the basis of mental state information.
The findings by \citet{strachan2024} also show that LLMs perform similarly to humans on most tasks requiring the inference of mental states. However, they also highlight the importance of systematic testing to ensure a non-superficial comparison between humans and AI.
\citet{hu-etal-2023-fine} probe LLMs with PragMega \citetlanguageresource{floyd_pragmega_resource}, initially developed to test different dimensions of pragmatic reasoning abilities in humans. Covering different modalities (text, image, audio), the benchmark includes various tasks, such as deceit, humor, irony, and metaphor, and is formatted as a multiple-choice question answering (MCQA). \citet{hu-etal-2023-fine} report that models lag behind humans, especially for humour and irony. 
Similarly, \citet{sravanthi-etal-2024-pub} introduce PUB \citeplanguageresource{PUBdataset}, a large-scale benchmark covering 14 tasks across implicature, presupposition, reference, and deixis. The authors combine new and existing datasets (e.g., GRICE, \citealp{zheng-etal-2021-grice}; IMPRESS, \citealp{jeretic-etal-2020-natural}; FigQA \citep{liu-etal-2022-testing_figQA}). In the evaluation, they highlight large variance across pragmatic phenomena and persistent gaps in performance between humans and LLMs. At the same time, new methodologies move beyond accuracy-based MCQA. A recent resource in this direction is PragmaticQA \citep{qi-etal-2023-pragmaticqa}, which targets open-domain open-ended pragmatic question answering, showing persistent struggles of state-of-the-art systems. Along the same lines, \citet{wu-etal-2024-rethinking} critique rigid multiple-choice evaluations and instead propose preference optimization with free-form evaluation protocols, where pragmatic quality is judged by human- or model-based raters across dimensions such as appropriateness and insightfulness. This connects pragmatic competence to deeper model representations, suggesting analogies to human high-level cognition.

\textbf{Non-English Benchmarks}
The above datasets have all been developed for English; hence, the findings are valid only in that setting. The performance of LLMs in pragmatic understanding for other languages remains an under-researched topic. \citet{park-etal-2024-multiprageval} propose MultiPragEval, extending an initially Korean pragmatics understanding benchmark to Chinese, German, and English via machine translation and post-editing. The dataset consists of potential violations of "conversational maxims" \citep{grice1975}. Their evaluation shows relatively good performance by closed-source LLMs, whereas open-source models perform far worse. They observe varying performance across languages, models, and the type of maxim violated.
Another fully native resource, SwordsmanImp \cite{yue2024largelanguagemodelsunderstand}, was compiled from Chinese sitcoms to evaluate conversational implicature. For European languages (other than the above-mentioned German), evaluations of pragmatic understanding are mostly entailed in more generic natural language understanding/inference benchmarks, or not addressed at all. Our motivation is therefore to address this gap, but also the share the experience gained with the non-trivial cultural adaptation of two nuanced language datasets.

\begin{figure*}[ht]
 \centering
 \begin{subfigure}{0.5\linewidth}
 \centering
 \includegraphics[width=0.8\linewidth]{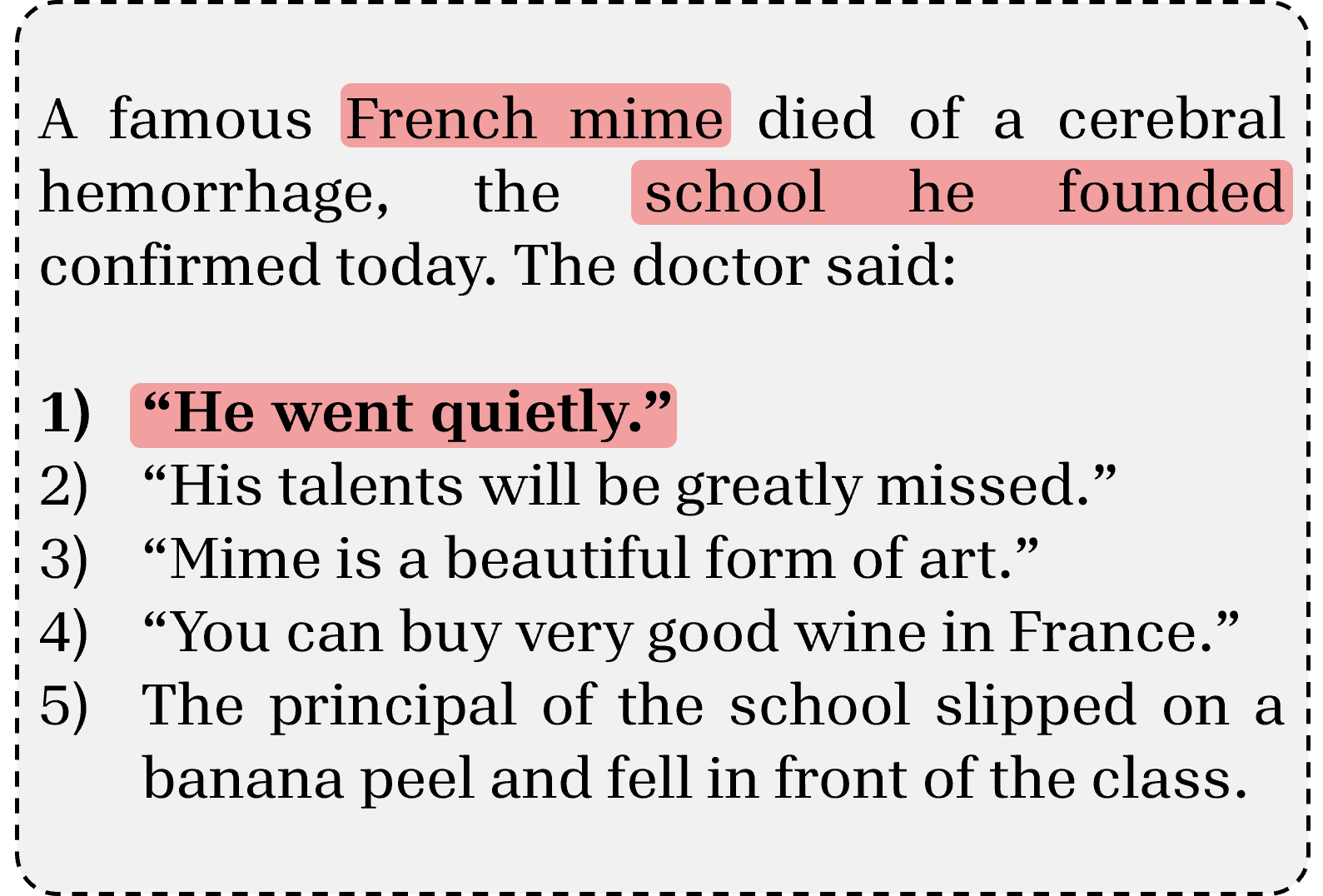}
 \caption{}\label{fig:ex_2a}
 \end{subfigure}\begin{subfigure}{0.5\linewidth}
 \centering
 \includegraphics[width=0.8\linewidth]{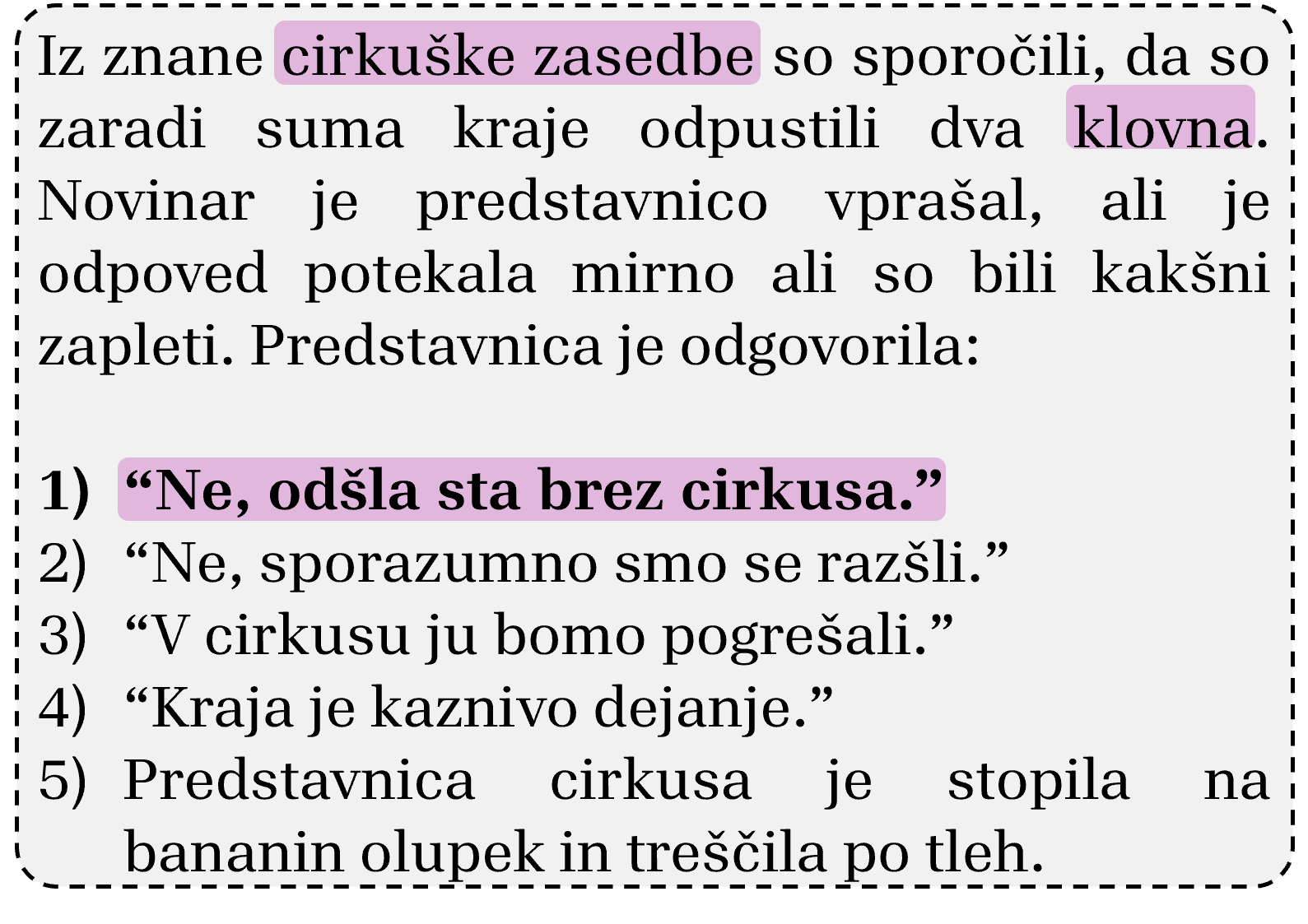}
 \caption{}\label{fig:ex_2b}
 \end{subfigure}
 \caption{Example from (Slo)PragMega: example from the Humor task. Original text on the left (a), Slovene example on the right (b). Problematic parts in red, adaptations in violet. Correct answer in bold.}
 \label{fig:example_2humour}
\end{figure*}

\begin{figure*}[ht!]
 \centering
 \begin{subfigure}{0.5\linewidth}
 \centering
 \includegraphics[width=0.8\linewidth]{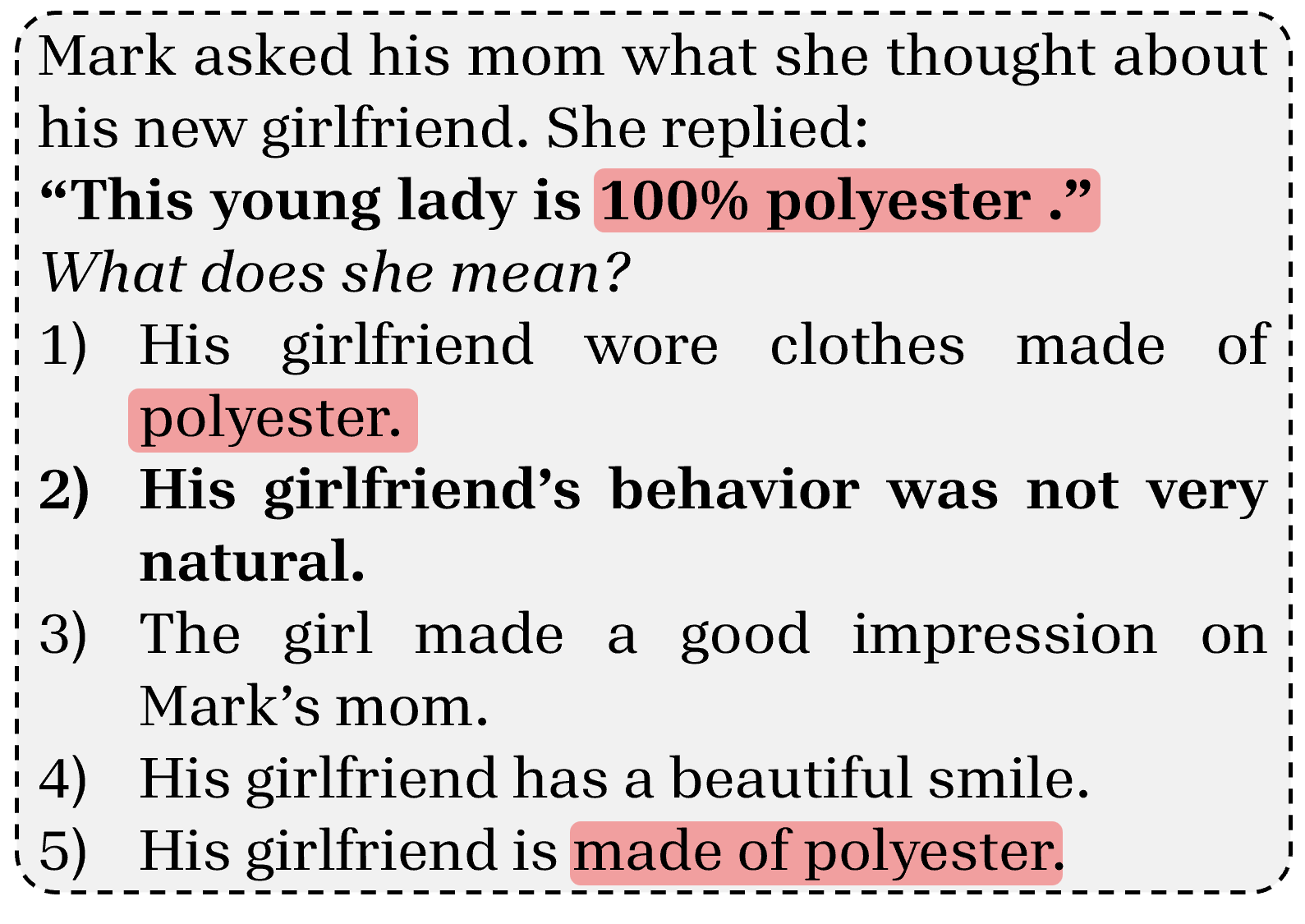}
 \caption{}
 \label{fig:ex_3a}
 \end{subfigure}\begin{subfigure}{0.5\linewidth}
 \centering
 \includegraphics[width=0.8\linewidth]{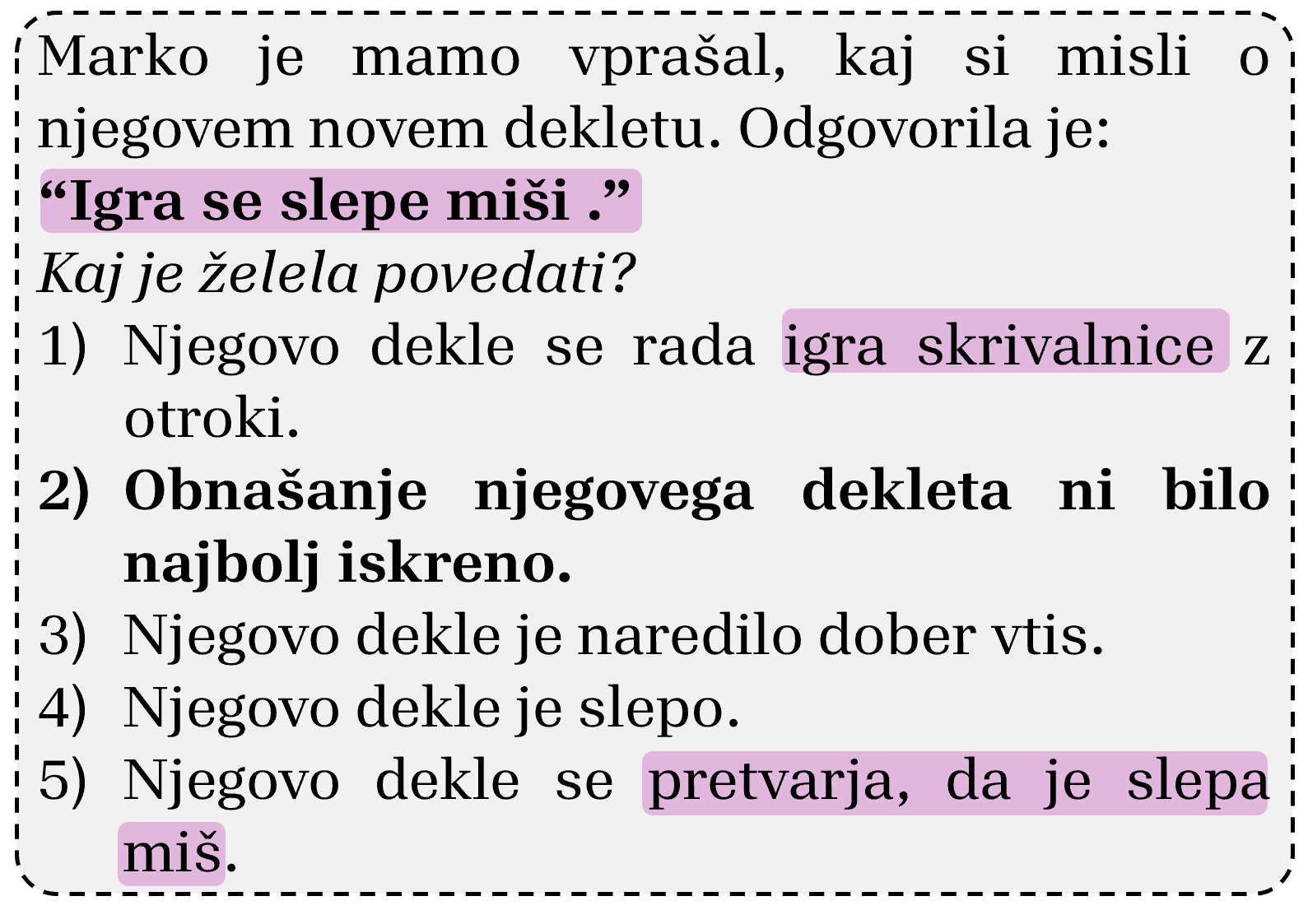}
 \caption{}
 \label{fig:ex_3b}
 \end{subfigure}
 \caption{Example from (Slo)PragMega: example from the Metaphor task. Original text on the left (a), Slovene example on the right (b). Problematic parts in red, adaptations in violet. Correct answer in bold.}
 \label{fig:metaphor_ex}
\end{figure*}

\section{Datasets}

While addressing different pragmatics phenomena, the two datasets presented here have a similar multiple-choice question answering (MCQA) format, as laid out by the original English datasets. 

Each example first describes a \textbf{Scenario} which provides an everyday situation with the context needed to resolve the pragmatic task (such as the participants, the setting, previous events, hints of emotional states). 
In the majority of the examples, the task is to discern the implied meaning of a speaker \textbf{Utterance} found at the end of the Scenario. 
Thus, the scenario is (usually) followed by a \textbf{Question} serving as the task instruction, e.g., \textit{What does PERSON mean?}. Then, four or five possible \textbf{Hypotheses} are provided as possible answers to the question.\footnote{The only exception to this is the Humour task in SloPragMega, in which the task setting and goal are somewhat different. Here, the initial Scenario does not include a speaker utterance, and no question directly follows the scenario. Rather, the task is to select the \textbf{Punchline} from the Hypotheses that create a joke or humorous effect by following from the initial \textbf{Situation}.} 

We describe the two datasets in further detail below.

\paragraph{SloPragEval} is the Slovene translation and adaptation of the MultiPragEval benchmark dataset \citep{park-etal-2024-multiprageval}, which was developed for the evaluation of LLMs on understanding speaker utterances that potentially flout one of the four Gricean maxims \citep{grice1975}: Quality, Quantity, Relevance, Manner, or those that do not (Literal utterances). The original benchmark consists of 300 task instances in four languages (Korean, German, English, and Chinese). The task instances are equally distributed between five categories: Quality, Quantity, Relevance, Manner, and Literal, and between the five answer options (A, B, C, D, E). 

We primarily opt to translate the English version to create examples in Slovene; however, as we describe in Section \ref{translating} below, other language versions were also consulted via machine translation for clarification, as the English version was insufficiently linguistically/culturally adapted. An example from the original dataset and its adaptation to Slovene is given in Figure \ref{fig:example_1MPE}.

Following recent considerations in benchmarking generative LLMs, especially the mitigation of contamination risks (see, e.g., \citealp[]{jacovi-etal-2023-stop}), we only publicly publish 60 examples (20\%) in totality, i.e., as labeled examples for development purposes, while the testing data (240 examples or 80\%) is provided without labels. 

\paragraph{SloPragMega} is a translation and adaptation of a section of the PragMega dataset \citeplanguageresource{floyd_pragmega_resource}, which was previously used for evaluation of LLMs in English by \citet{hu-etal-2023-fine} and \citet{wu-etal-2024-rethinking}. The original resource was constructed to cover 20 tasks, spanning 11 phenomena (e.g., indirect speech, irony, scalar implicatures). The dataset was manually crafted by psychologists and aimed at discovering whether "pragmatic inferencing" is a result of a single cognitive skill or, on the contrary, of different dissociable skills depending on the type of phenomena encountered.

While all of the phenomena in the dataset are relevant for pragmatics understanding and evaluation, not all of them are are at the same level of difficulty\footnote{For example, the "Coherence" task is very similar to natural language entailment tasks, as it only consists of two sentences, where the other is either coherent with the first one or not, the resolution of which usually rests on world knowledge.}. Secondly, many of these phenomena may overlap with the SloPragEval examples\footnote{"Indirect requests", "Conversational implicatures", "Irony", "Metaphor" can all be conveyed via maxim-flouting utterances.}.

To create the first Slovene version of the dataset, we thus only select three tasks: Irony, Metaphor, and Humour. These consist of 50, 30, and 25 examples, respectively, or 105 examples in total. We provide two examples from the original dataset and its adaptations to Slovene in Figure \ref{fig:example_2humour} (Humour task) and Figure \ref{fig:metaphor_ex} (Metaphor task). 

Due to the smaller size of the dataset, we only publish 5 examples (approx. 5\%) as labeled data for development, and provide the remaining examples (100, 95\%) as unlabelled test data. Compared to the original dataset, we shuffle the responses to ensure that the answer types (e.g., literal meaning, metaphorical meaning, distractor) appear in different positions (1-5), and that the correct answers are evenly distributed across these positions.

Both datasets are available on the Slovene benchmarking platform SloBench\footnote{\url{https://slobench.cjvt.si/}}.
We report the benchmark dataset sizes and splits in Table \ref{tab:datasets}.

\begin{table}[htb]
 \centering
 \begin{tabular}{c|c|c}
 \textbf{Dataset} & \textbf{test} &  \textbf{dev} \\
 \hline
 SloPragEval & 240 & 60 \\
 SloPragMega & 100 & 5 \\
 \end{tabular}
 \caption{Benchmark dataset sizes}
 \label{tab:datasets}
\end{table}


\paragraph{Translation}\label{translating}

Several steps were carried out to translate and adapt the dataset from English to Slovene.

Initially, the first step of translating the dataset was to recruit students enrolled in MA Translation Studies and MA Digital Linguistics at the Faculty of Arts, University of Ljubljana. The student project involved translation and peer revision, with multiple rounds of discussions and online voting for proposed solutions.
Finally, after the student translation and revision stages, several rounds of revision were performed by two expert linguists and translators (authors of this article). Additionally, some minor corrections were also suggested through the crowdsourcing campaign (see Section \ref{sec:humanbaseline})

\paragraph{Localization Challenges}

For most tasks, translation was far from straightforward. Rather, the task examples from both datasets had to be considerably altered and adapted to the target linguistic and sociocultural context. The alterations ranged from minor linguistic and cultural adaptations (e.g., exchanging the idiomatic phrase in the utterance, or localizing proper names) to complete substitution (e.g., a non-translatable pun-based joke). We categorize these adaptations into two classes and provide examples.
First are various \textit{linguistic challenges} common to translation, which encompass everything from differences in syntax, semantics, pragmatics, and text stylistics. Cases that demanded thorough adaptation to produce natural-sounding language were idioms, metaphors, fixed phrases, puns, homonyms, ambiguities, and genre conventions. 
Secondly, the texts included many \textit{cultural specifics} such as geographical names, person names, and typical culture-bound concepts (food, clothes, holidays, flora, fauna, law, architecture etc.). As is demonstrated by the example in Figure \ref{fig:example_1MPE}, the English source text contains many culturally specific items such as names \textit{Emily}, \textit{United States}, but also the culture-bound concept of a \textit{forest resort}. All of these had to be replaced with more suitable and familiar equivalents, for example, \textit{forest resort} became a \textit{koča} 'cabin'. The utterance and its intended meaning, however, were kept unchanged.

Moreover, we noticed that even in the English source texts from MultiPragEval, which had previously been translated from Korean, many of these were not sufficiently adapted and sometimes impeded understanding. The translators and/or reviewers had to consult the text in other language variants by using machine translation to uncover the intended meaning, find the relation between the utterance and answer hypotheses, or clarify ambiguous phrasing. In several cases, this revealed issues in the source material itself that had not been adequately transferred ("translationese", e.g., phrases that demonstrate "shining through"; cultural mismatches; and cases where the utterance itself had been adequately adapted, but not the answer hypotheses).

The greatest challenges were most markedly present in translating the examples from the Humour task. Here, both situational and linguistic elements highly influence the understanding of the joke. For example, puns can rest on common linguistic phenomena such as polysemy or homonymy, where the multiple possible resolutions create a certain incongruency or opposition \citep{Attardo_2010,ATTARDORASKIN+1991+293+348}. An example from the Humour task, which had to be considerably modified, is depicted in Figure \ref{fig:example_2humour}. The English situation contains elements that may be unfamiliar to Slovene readers, as they are relatively rare in the target culture (\textit{French mime}, \textit{a school founded by an individual}). Secondly, the phrase \textit{go quietly} in the punchline carries multiple meanings ('without noise' and 'peacefully'), which allows it to function in those two conflicting contexts (and thus creating the joke). Its literal translation (\textit{potiho} 'quietly') does not have the same semantic profile.
To adapt it into an equivalent example in Slovene, we considered the original scenario and tried to find another phrase that would relate to a similar scenario and carry two (sufficiently different) meanings. The solution was to use the phrase \textit{brez cirkusa} 'without [the] circus', which can also be used metaphorically in the sense 'without making a fuss'. This then led us to change the initial situation, which now concerns a renowned circus band that dismissed two clowns on suspicion of theft.

\section{Human Baseline Campaign}\label{sec:humanbaseline}

Following the construction of the larger pragmatics understanding dataset SloPragEval, we conducted a crowdsourcing campaign to administer the dataset to human annotators. The goal of this external validation was two-fold: first, to validate the dataset itself in terms of general intelligibility, and, second, to create a human baseline against which we can later compare the performance of language models.

To recruit annotators, we organized a crowdsourcing campaign via various social media channels for annotators to apply for participation. To avoid any data leakage, we only distributed the tests via direct e-mails to participants, who were instructed to upload their solutions anonymously to a private cloud. 
Due to the size of the pragmatics test, we opted to split the dataset into smaller chunks to still obtain volunteer annotators. Thus, each of the annotators was sent 50 randomly selected examples. Since the pragmatics understanding task was self-explanatory in the sense that each example already contained the task question (\textit{What did <PERSON> mean?}), the annotators were not given any additional instructions or clarifications about the underlying data and task.

In total, 79 questionnaires were sent out, and 57 complete responses were received from the informants. This resulted in a minimum of 6 human answers for each of the 300 examples in the dataset.

To compute a human baseline, we first calculated the accuracy per rater on the 50-item questionnaire (Human-IND). Then, we also accumulated the individual raters' responses into six complete sets of human responses, and calculated the accuracy for each of the six sets (Human-SET)\footnote{On the 240 items from the test split only.}. 
The human baseline is reported in Table \ref{pragEval_results_SL}. We observe that both the average accuracies, i.e., by individual raters (Human-IND) or on aggregated answers, are around 0.85. Secondly, we observe that humans do not perform on the same level across all maxim violations: it seems Manner-violating utterances were the most difficult to interpret, with accuracies as low as 0.67. On the other hand, Literal utterances are more readily comprehensible, with average accuracy over 0.90. Moreover, we can also observe considerable variation in performance between individual raters (Human-IND), with standard deviations as large as 0.16 in the case of Manner. 


\begin{table*}[tb!]\centering
\begin{tabular}{l|ccc|c}\centering
\textbf{Model}  & \textbf{Metaphor}   & \textbf{Irony} & \textbf{Humour} & \textbf{Average} \\
\hline
DS-DQ-14B & 0.67 ±0.04 & 0.74 ±0.10 & 0.54 ±0.04  & 0.65   ±0.06 \\
Gemma3-27B   & \textit{\textbf{0.89}} ±0.00   & \textbf{\textit{0.94}} ±0.00 & \textbf{0.78} ±0.02 & \textbf{\textit{0.87}} ±0.01 \\
GaMS-27B & 0.87 ±0.05 & 0.81 ±0.02  & 0.42 ±0.07  & 0.70 ±0.04  \\
Llama3.3-70B  & \textit{\textbf{0.89}} ±0.00 & 0.85 ±0.04 & 0.68 ±0.05   & 0.81 ±0.01 \\
\hline
GPT-5-chat & 0.96 ± 0.00	&0.94 ± 0.00	&0.89 ± 0.02	&0.93 ± 0.01 \\
GPT-5 &\textbf{1.00} ± 0.00	&\textbf{0.96} ± 0.02	&\textbf{1.00} ± 0.00	&\textbf{0.99} ± 0.01 \\
\hline
\end{tabular}
\caption{Accuracy scores on SloPragMega, prompting in Slovene. Best score per phenomenon in bold, best score per phenomenon among open source models in bold italic.  }\label{pragmega_results_SL}
\end{table*}

\begin{table*}[tb!]\centering
\begin{tabular}{l|ccc|c}
\textbf{Model}  & \textbf{Metaphor} & \textbf{Irony}   & \textbf{Humour}  & \textbf{Average}  \\
\hline
DS-DQ-14B  & 0.68 ±0.04 & 0.72 ±0.06   & 0.50 ±0.04 & 0.63   ±0.04 \\
Gemma3-27B  & \textbf{\textit{0.93}} ±0.00 & \textbf{\textit{0.94}} ±0.00 & 0.68 ±0.02  & \textit{\textbf{0.85}} ±0.01\\
GaMS-27B & 0.81 ±0.02  & 0.82 ±0.01  & 0.43 ±0.02   & 0.69 ±0.02  \\   
Llama3.3-70B & 0.88 ±0.02 & 0.85 ±0.02 & \textbf{\textit{0.69}} ±0.02 & 0.81 ±0.01  \\
\hline
GPT-5-chat &0.96 ± 0.00	&0.94 ± 0.00	&\textbf{1.00} ± 0.00	&0.97 ± 0.00 \\
GPT-5 & \textbf{1.00 }± 0.00	&\textbf{0.97} ± 0.01	&\textbf{1.00} ± 0.00	&\textbf{0.99} ± 0.00\\
\hline
\end{tabular}
\caption{Accuracy scores on SloPragMega, prompting in English. Best score per phenomenon in bold, best score per phenomenon among open source models in bold italic. }\label{pragmega_results_EN}
\end{table*}

\section{LLM Evaluation}

To evaluate LLMs on pragmatics understanding, we separately administer the SloPragMega and SloPragEval testing data, which consist of 100 and 240 examples, respectively. 
Following previous research, we administer the test in an MCQA setting. As the latter can be considered and evaluated as a classification task, we use the traditional Accuracy metric to quantify performance. However, given the non-deterministic nature of generative LLMs, we collect predictions\footnote{We extract the single-letter/single-digit answers using regular expression matching and manually check for and correct irregularities.} and average the results from multiple (3) test runs, keeping the default model settings such as temperature. We provide further details about the models used and task prompts in the following subsections.

\subsection{Models}
We evaluate instruction-tuned generative models, including four locally installed open source models and two closed-source models. The open-source models\footnote{All the open-source models are 4-bit quantized.} include the 14B version of DeepSeek-R1-Distil-Qwen (DS-DQ-14B, \citealp{deepseekai2025deepseekr1incentivizingreasoningcapability})\footnote{\url{https://huggingface.co/deepseek-ai/DeepSeek-R1-Distill-Qwen-14B}}, the 27B version of Gemma 3 \citep{gemmateam2025gemma3technicalreport}\footnote{\url{https://huggingface.co/google/gemma-3-27b-it}}, and the 70B version of Llamma 3.3\footnote{\url{https://huggingface.co/meta-llama/Llama-3.3-70B-Instruct}}, which have multilingual support. Furthermore, we also evaluate the 27B version of GaMS\footnote{\url{https://huggingface.co/cjvt/GaMS-27B-Instruct/}}, a Slovene generative model based on Google's Gemma 2 \citep{gemmateam2024gemma2improvingopen} family and continually pretrained on Slovene and English, and partially also on Croatian, Serbian, and Bosnian.
The two closed-source models we use are OpenAI's GPT-5\footnote{API name: \textit{gpt-5-2025-08-07}, last update 2025-08-01.} and GPT-5-chat \citep{openAIgpt5}\footnote{API name: \textit{gpt-5-chat-latest}, last update 2025-08-01.}, which we access via their proprietary API\footnote{https://platform.openai.com/docs/api-reference/}. 

\subsection{SloPragEval}
To evaluate LLMs on SloPragEval, we follow the original strategy used by \citet{park-etal-2024-multiprageval} without any additional information. That is, the complete prompt to the model directly starts with the example task: the Scenario and Utterance, the task Question\footnote{\textit{Med naslednjimi možnostmi izberi najustreznejši pomen zgornje izjave:} / \textit{Choose the most appropriate meaning of the above utterance from the following options.}}, and the answer Hypotheses.


\subsection{SloPragMega}
To evaluate LLMs on SloPragMega, we follow the original prompts used by \citet{hu-etal-2023-fine} (available in the Appendix). 
These consist of a short task description, the situation, and the answer hypotheses. We test English and Slovene variants of the same prompt format.

\section{Results}

The results on the two datasets indicate that the models have advanced in understanding more nuanced utterances.

\begin{table*}[tb]
\small
\centering
\begin{tabular}{l|ccccc|c}
\textbf{Agent } & \textbf{Quality}	& \textbf{Quantity}	& \textbf{Relation}	& \textbf{Manner}	& \textbf{Literal}	& \textbf{Average} \\
\hline
Human-IND & 0.90 ± 0.09	& 0.84 ± 0.12	& 0.86 ± 0.14	& 0.68 ± 0.16	& 0.93 ± 0.09	& 0.84 ± 0.06\\
Human-SET & 0.92 ± 0.02	& 0.81 ± 0.09	& 0.89 ± 0.04	& 0.67 ± 0.03	& 0.95 ± 0.05	& 0.85 ± 0.03 \\
\hline \hline
DS-DQ-14B& 0.27 ± 0.04 & 0.31 ± 0.04 & 0.44 ± 0.08 & 0.33 ± 0.07 & 0.81 ± 0.05 & 0.43   ± 0.04 \\
Gemma3-27B & \textbf{\textit{0.83}} ± 0.02 & 0.57 ± 0.01 & \textbf{\textit{0.82}} ± 0.01 & 0.59 ± 0.01 & 0.96 ± 0.00 & 0.75 ± 0.01 \\
GaMS-27B & 0.64 ± 0.08 & 0.50 ± 0.00 & 0.69 ± 0.04 & 0.56 ± 0.06 & 0.85 ± 0.02 & 0.65 ± 0.02 \\
Llama3.3-70B & 0.81 ± 0.00 & \textbf{\textit{0.64}} ± 0.03 & \textbf{\textit{0.82}} ± 0.02 & \textbf{\textit{0.62}} ± 0.01 & \textbf{{0.98}} ± 0.00 & \textbf{\textit{0.77}} ± 0.01 \\
\hline
GPT-5-chat &0.88 ± 0.02 & \textbf{0.76} ± 0.02 & \textbf{0.86} ± 0.03 & 0.61 ± 0.01 & 0.94 ± 0.01 & \textbf{0.81} ± 0.01 \\
GPT-5 & \textbf{0.92} ± 0.01 & 0.66 ± 0.03 & 0.85 ± 0.03 & \textbf{0.67} ± 0.06 & 0.97 ± 0.01 & \textbf{0.81} ± 0.02\\
\hline
\end{tabular}
\caption{Accuracy scores on SloPragEval, prompting in Slovene. Human baseline reported per individual rater (Human-IND) and per aggregated set (Human-SET). Best model per phenomenon in bold, best score per phenomenon among open source models in bold italic. }\label{pragEval_results_SL}
\end{table*}

\begin{table*}[tb]
\small
\centering
\begin{tabular}{l|ccccc|c}
\textbf{Agent } & \textbf{Quality}	& \textbf{Quantity}	& \textbf{Relation}	& \textbf{Manner}	& \textbf{Literal}	& \textbf{Average} \\
\hline
DS-DQ-14B & 0.44 ± 0.06	& 0.42 ± 0.02	& 0.52 ± 0.04	& 0.41 ± 0.06	& 0.78 ± 0.02	& 0.51 ± 0.02\\
Gemma3-27B	&0.78 ± 0.01	&0.55 ± 0.01	&0.81 ± 0.01	&0.59 ± 0.01	&\textbf{{0.98}} ± 0.00	&0.74 ± 0.00\\
GaMS-27B   & 0.56 ± 0.02	&0.48 ± 0.04	&0.48 ± 0.04	&0.51 ± 0.08	&0.81 ± 0.06	&0.57 ± 0.01\\
Llama3.3-70B & \textbf{\textit{0.82}} ± 0.01	&\textbf{\textit{0.67}} ± 0.03	&\textbf{\textit{0.85}} ± 0.04&	\textbf{\textit{0.61}} ± 0.03&	\textbf{0.98} ± 0.00	& \textbf{\textit{0.79}} ± 0.01\\
\hline
GPT-5-chat &\textbf{0.92} ± 0.02	& \textbf{0.70} ± 0.01	& \textbf{0.90} ± 0.02	& {0.67} ± 0.00	& 0.94 ± 0.01	& \textbf{0.83} ± 0.01 \\
GPT-5  & 0.89 ± 0.01 & 0.69 ± 0.04	& 0.85 ± 0.03	& \textbf{0.68} ± 0.01	& \textbf{0.98} ± 0.00 & 0.82 ± 0.01\\
\hline
\end{tabular}
\caption{Accuracy scores on SloPragEval, prompting in English. Best model per phenomenon in bold, best score per phenomenon among open source models in bold italic. }\label{pragEval_results_EN}
\end{table*}

Considering first the results on the smaller SloPragMega benchmark in Table \ref{pragmega_results_SL} (using Slovene prompts) and Table \ref{pragmega_results_EN} (using English prompts), the closed models already achieve perfect scores on some tasks. For example, while smaller open source models still struggle quite a bit with resolving the tasks, especially in selecting Punchlines in the Humour task when prompted in Slovene (e.g., accuracies ranging from 0.42 for GaMS to 0.78 for Gemma), GPT-5 achieves a whopping 1.00 accuracy. 
We also note that model size does not necessarily translate to better performance among open-source models. While we observe differences between the 14B DeepSeek-R1-Distil-Qwen and other larger models, there are no significant differences in performance between the two 27B models and the 70B Llamma 3.3. In fact, the smaller Gemma 3 outperforms its larger rival on many occasions. Relating to the language of the prompt, the models seem to perform very similarly or even better when the task descriptions and the question are in Slovene. 

Results on SloPragEval (Table \ref{pragEval_results_SL} and Table \ref{pragEval_results_EN}), however, show a more complex picture. Several observations can be made: while two of the open-source models are still relatively far from the human baseline (lowest average score of 0.43/0.51 using Slovene/English prompt vs. the human baseline of 0.85), the state-of-the-art GPT-5 achieves accuracy (0.81/0.83 using Slovene/English prompt) that is practically on par with human performance.

However, differences may be observed in performance by utterance type. Humans and LLMs have no difficulties in understanding Literal utterances. Violations of Quality (e.g., metaphors, irony), Relation (stating not directly relevant facts), and Quantity (saying less/more than expected) are also largely comprehensible by humans. Manner-flouting utterances seem to be a difficult task for both humans and LLMs: here, humans and best-performing LLMs only achieve an accuracy of 0.68, whereas smaller open-source models achieve scores as low as 0.33/0.41 following a Slovene/English prompt.

The biggest gap between human and LLM performance can be observed in the Quantity category. Humans can correctly interpret over 80\% of Quantity-flouting utterances, while the best LLM correctly interprets 76\% (GPT when prompted in Slovene). The open-source model scores are much lower and range from 0.31-0.64 when prompted in Slovene, and 0.42-0.67 when prompted and English. Contrary to the results on the SloPragMega dataset, the models seem to perform similarly or slightly better on examples from SloPragEval when prompted in English.

Comparing these results to those reported by \citet{park-etal-2024-multiprageval} for English, Korean, German, and Chinese, some additional observations can be made. Back in 2024, the best-performing proprietary model for English was Claude3-Opus, reaching 0.85 accuracy, and an even higher 0.87 score for Korean, while GPT-4 achieved 0.75 for English and 0.81 for Korean. Interestingly, proprietary models performed better for Korean than for English. It would appear that the average score for Slovene with GPT-5 is on par with the one by GPT-4 for Korean, perhaps indicating that pragmatic understanding has not dramatically improved between these two models. We also observe a similar phenomenon regarding the task types: most models perform the worst on Manner-flouting utterances.


\section{Conclusion}

We have presented two new benchmark datasets for Slovene, SloPragMega and SloPragEval, to evaluate the understanding of nuanced language, which requires mastering all levels of language and also the knowledge of the social and cultural context.

We have first demonstrated the challenges that arise in creating such datasets through the translation of established resources. In the process, we have stumbled upon many instances that prevented simple translation or adaptation but demanded complete rescripts.
The results of the evaluation of LLMs show that, on average, LLMs are reaching or have already reached human performance in understanding various pragmatic phenomena. However, this stands mostly for the best-performing closed-source models, while smaller open-source models somewhat lag behind. 

The high performance might be due to several factors. First, despite many adaptations, large overlaps with the source texts still exist, which allow the LLMs to potentially pivot via English. Secondly, we cannot prohibit the possibility of contamination, meaning the models have already seen the underlying original datasets. We thus argue that future endeavours of benchmark creation should strive for bottom-up approaches, which would lead to more linguistically and culturally genuine contexts as well as more difficult examples.

\section*{Acknowledgements}

This research was supported by the Slovene Research and Innovation Agency (ARRS/ARIS) through the project \textit{Large Language Models for Digital Humanities}, grant number GC-0002, and through the research programme \textit{Slovene Language - Basic, Contrastive, and Applied Studies}, grant number P6-0215.

\section*{Limitations}
Our initial experiments with LLMs feature only a small set of models. Future evaluations should include more models, both in terms of provenance and size. For instance, the 27B GaMS model is, according to the developers, still undertrained for Slovene, so a bigger 100B version that is under construction could provide much better results.
Secondly, we concur with other researchers who argue for open-ended evaluations as the more natural choice for generative models; however, we reserve such evaluations for further studies. We have also not carried out detailed evaluations of the generated response, which frequently contained the reasoning behind the selection and explanations of the phenomena at hand. We plan to address this in the future, as it could provide a further glimpse into the language understanding capabilities of LLMs.

\section{Bibliographical References}\label{sec:reference}

\bibliographystyle{lrec2026-natbib}
\bibliography{bibliography}

\begin{thebibliography}{28}
\expandafter\ifx\csname natexlab\endcsname\relax\def\natexlab#1{#1}\fi

\bibitem[{AlKhamissi et~al.(2024)AlKhamissi, ElNokrashy, Alkhamissi, and Diab}]{alkhamissi-etal-2024-investigating}
Badr AlKhamissi, Muhammad ElNokrashy, Mai Alkhamissi, and Mona Diab. 2024.
\newblock \href {https://doi.org/10.18653/v1/2024.acl-long.671} {Investigating cultural alignment of large language models}.
\newblock In \emph{Proceedings of the 62nd Annual Meeting of the Association for Computational Linguistics (Volume 1: Long Papers)}, pages 12404--12422, Bangkok, Thailand. Association for Computational Linguistics.

\bibitem[{Attardo(2010)}]{Attardo_2010}
Salvatore Attardo. 2010.
\newblock \href {https://doi.org/10.1515/9783110219029} {\emph{Linguistic Theories of Humor}}, volume~1.
\newblock De Gruyter.

\bibitem[{Attardo and Raskin(1991)}]{ATTARDORASKIN+1991+293+348}
Salvatore Attardo and Victor Raskin. 1991.
\newblock \href {https://doi.org/doi:10.1515/humr.1991.4.3-4.293} {Script theory revis(it)ed: joke similarity and joke representation model}.
\newblock \emph{HUMOR}, 4(3-4):293--348.

\bibitem[{Birner(2012)}]{birner_pragmatics_12}
Betty~J. Birner. 2012.
\newblock \emph{Introduction to Pragmatics}, 1st edition.
\newblock Wiley Publishing.

\bibitem[{Boisson et~al.(2025)Boisson, Siddique, Borkakoty, Antypas, Espinosa~Anke, and Camacho-Collados}]{boisson-etal-2025-automatic}
Joanne Boisson, Zara Siddique, Hsuvas Borkakoty, Dimosthenis Antypas, Luis Espinosa~Anke, and Jose Camacho-Collados. 2025.
\newblock \href {https://aclanthology.org/2025.coling-main.448/} {Automatic extraction of metaphoric analogies from literary texts: Task formulation, dataset construction, and evaluation}.
\newblock In \emph{Proceedings of the 31st International Conference on Computational Linguistics}, pages 6692--6704, Abu Dhabi, UAE. Association for Computational Linguistics.

\bibitem[{Choi et~al.(2023)Choi, Pei, Kumar, Shu, and Jurgens}]{choi-etal-2023-llms}
Minje Choi, Jiaxin Pei, Sagar Kumar, Chang Shu, and David Jurgens. 2023.
\newblock \href {https://doi.org/10.18653/v1/2023.emnlp-main.699} {Do {LLM}s understand social knowledge? evaluating the sociability of large language models with {S}oc{KET} benchmark}.
\newblock In \emph{Proceedings of the 2023 Conference on Empirical Methods in Natural Language Processing}, pages 11370--11403, Singapore. Association for Computational Linguistics.

\bibitem[{{DeepSeek-AI}(2025)}]{deepseekai2025deepseekr1incentivizingreasoningcapability}
{DeepSeek-AI}. 2025.
\newblock \href {http://arxiv.org/abs/2501.12948} {Deepseek-r1: Incentivizing reasoning capability in llms via reinforcement learning}.

\bibitem[{{Gemma Team}(2024)}]{gemmateam2024gemma2improvingopen}
{Gemma Team}. 2024.
\newblock \href {http://arxiv.org/abs/2408.00118} {Gemma 2: Improving open language models at a practical size}.

\bibitem[{{Gemma Team}(2025)}]{gemmateam2025gemma3technicalreport}
{Gemma Team}. 2025.
\newblock \href {http://arxiv.org/abs/2503.19786} {Gemma 3 technical report}.

\bibitem[{Grice(1975)}]{grice1975}
H.~P. Grice. 1975.
\newblock Logic and conversation.
\newblock In P.~Cole and J.~Morgan, editors, \emph{Syntax and Semantics}, volume~3, pages 22--40. Academic Press.
\newblock Reprinted as ch.2 of Grice 1989, 22–40.

\bibitem[{Halat and Atlamaz(2024)}]{halat-atlamaz-2024-implicatr}
Mustafa Halat and {\"U}mit Atlamaz. 2024.
\newblock \href {https://aclanthology.org/2024.sigturk-1.3/} {{I}mplica{TR}: A granular dataset for natural language inference and pragmatic reasoning in {T}urkish}.
\newblock In \emph{Proceedings of the First Workshop on Natural Language Processing for Turkic Languages (SIGTURK 2024)}, pages 29--41, Bangkok, Thailand and Online. Association for Computational Linguistics.

\bibitem[{Hu et~al.(2023)Hu, Floyd, Jouravlev, Fedorenko, and Gibson}]{hu-etal-2023-fine}
Jennifer Hu, Sammy Floyd, Olessia Jouravlev, Evelina Fedorenko, and Edward Gibson. 2023.
\newblock \href {https://doi.org/10.18653/v1/2023.acl-long.230} {A fine-grained comparison of pragmatic language understanding in humans and language models}.
\newblock In \emph{Proceedings of the 61st Annual Meeting of the Association for Computational Linguistics (Vol. 1: Long Papers)}, pages 4194--4213. Association for Computational Linguistics.

\bibitem[{Jacovi et~al.(2023)Jacovi, Caciularu, Goldman, and Goldberg}]{jacovi-etal-2023-stop}
Alon Jacovi, Avi Caciularu, Omer Goldman, and Yoav Goldberg. 2023.
\newblock \href {https://doi.org/10.18653/v1/2023.emnlp-main.308} {Stop uploading test data in plain text: Practical strategies for mitigating data contamination by evaluation benchmarks}.
\newblock In \emph{Proceedings of the 2023 Conference on Empirical Methods in Natural Language Processing}, pages 5075--5084, Singapore. Association for Computational Linguistics.

\bibitem[{Jeretic et~al.(2020)Jeretic, Warstadt, Bhooshan, and Williams}]{jeretic-etal-2020-natural}
Paloma Jeretic, Alex Warstadt, Suvrat Bhooshan, and Adina Williams. 2020.
\newblock \href {https://doi.org/10.18653/v1/2020.acl-main.768} {Are natural language inference models {IMPPRESsive}? {L}earning {IMPlicature} and {PRESupposition}}.
\newblock In \emph{Proceedings of the 58th Annual Meeting of the Association for Computational Linguistics}, pages 8690--8705, Online. Association for Computational Linguistics.

\bibitem[{Jones et~al.(2024)Jones, Trott, and Bergen}]{jones-etal-2024-comparing-humans}
Cameron~R. Jones, Sean Trott, and Benjamin Bergen. 2024.
\newblock \href {https://doi.org/10.1162/tacl_a_00674} {Comparing humans and large language models on an experimental protocol inventory for theory of mind evaluation ({EPITOME})}.
\newblock \emph{Transactions of the Association for Computational Linguistics}, 12:803--819.

\bibitem[{Liu et~al.(2022)Liu, Cui, Zheng, and Neubig}]{liu-etal-2022-testing_figQA}
Emmy Liu, Chenxuan Cui, Kenneth Zheng, and Graham Neubig. 2022.
\newblock \href {https://doi.org/10.18653/v1/2022.naacl-main.330} {Testing the ability of language models to interpret figurative language}.
\newblock In \emph{Proceedings of the 2022 Conference of the North American Chapter of the Association for Computational Linguistics: Human Language Technologies}, pages 4437--4452, Seattle, United States. Association for Computational Linguistics.

\bibitem[{{OpenAI}(2025)}]{openAIgpt5}
{OpenAI}. 2025.
\newblock Introducing {GPT}-5.
\newblock \emph{https://openai.com/index/introducing-gpt-5/}.

\bibitem[{Park et~al.(2024)Park, Lee, Park, Jeong, Koo, Hwang, Park, and Lee}]{park-etal-2024-multiprageval}
Dojun Park, Jiwoo Lee, Seohyun Park, Hyeyun Jeong, Youngeun Koo, Soonha Hwang, Seonwoo Park, and Sungeun Lee. 2024.
\newblock \href {https://doi.org/10.18653/v1/2024.genbench-1.7} {{M}ulti{P}rag{E}val: Multilingual pragmatic evaluation of large language models}.
\newblock In \emph{Proceedings of the 2nd GenBench Workshop on Generalisation (Benchmarking) in NLP}, pages 96--119. Association for Computational Linguistics.

\bibitem[{Qi et~al.(2023)Qi, Du, Manning, and Huang}]{qi-etal-2023-pragmaticqa}
Peng Qi, Nina Du, Christopher Manning, and Jing Huang. 2023.
\newblock \href {https://doi.org/10.18653/v1/2023.findings-acl.385} {{P}ragmati{CQA}: A dataset for pragmatic question answering in conversations}.
\newblock In \emph{Findings of the Association for Computational Linguistics: ACL 2023}, pages 6175--6191, Toronto, Canada. Association for Computational Linguistics.

\bibitem[{Qu and Wang(2024)}]{yao_jue_performance_biases_24}
Yao Qu and Jue Wang. 2024.
\newblock \href {https://doi.org/10.1057/s41599-024-03609-x} {Performance and biases of large language models in public opinion simulation}.
\newblock \emph{Humanities and Social Sciences Communications}, 11.

\bibitem[{Sileo et~al.(2022)Sileo, Muller, Van~de Cruys, and Pradel}]{sileo2022pragmaticscenteredevaluationframeworknatural}
Damien Sileo, Philippe Muller, Tim Van~de Cruys, and Camille Pradel. 2022.
\newblock \href {https://aclanthology.org/2022.lrec-1.255/} {A pragmatics-centered evaluation framework for natural language understanding}.
\newblock In \emph{Proceedings of the Thirteenth Language Resources and Evaluation Conference}, pages 2382--2394, Marseille, France. European Language Resources Association.

\bibitem[{Sravanthi et~al.(2024)Sravanthi, Doshi, Tankala, Murthy, Dabre, and Bhattacharyya}]{sravanthi-etal-2024-pub}
Settaluri Sravanthi, Meet Doshi, Pavan Tankala, Rudra Murthy, Raj Dabre, and Pushpak Bhattacharyya. 2024.
\newblock \href {https://doi.org/10.18653/v1/2024.findings-acl.719} {{PUB}: A pragmatics understanding benchmark for assessing {LLM}s' pragmatics capabilities}.
\newblock In \emph{Findings of the Association for Computational Linguistics: ACL 2024}, pages 12075--12097. Association for Computational Linguistics.

\bibitem[{Strachan et~al.(2024)Strachan, Albergo, Borghini, Pansardi, Scaliti, Gupta, Saxena, Rufo, Panzeri, Manzi, Graziano, and Becchio}]{strachan2024}
James W.~A. Strachan, Dalila Albergo, Giulia Borghini, Oriana Pansardi, Eugenio Scaliti, Saurabh Gupta, Krati Saxena, Alessandro Rufo, Stefano Panzeri, Guido Manzi, Michael S.~A. Graziano, and Cristina Becchio. 2024.
\newblock Testing theory of mind in large language models and humans.
\newblock \emph{Nature Human Behaviour}, 8:1285–1295.

\bibitem[{Wen and Tian(2025)}]{WenTian+2025+259+283}
Xu~Wen and Yaling Tian. 2025.
\newblock \href {https://doi.org/doi:10.1515/ip-2025-2004} {Understanding ironic utterances: A comprehensive examination of chatgpt-4o}.
\newblock \emph{Intercultural Pragmatics}, 22(2):259--283.

\bibitem[{Wu et~al.(2024)Wu, Yang, Chen, and Su}]{wu-etal-2024-rethinking}
Shengguang Wu, Shusheng Yang, Zhenglun Chen, and Qi~Su. 2024.
\newblock \href {https://doi.org/10.18653/v1/2024.emnlp-main.1258} {Rethinking pragmatics in large language models: Towards open-ended evaluation and preference tuning}.
\newblock In \emph{Proceedings of the 2024 Conference on Empirical Methods in Natural Language Processing}, pages 22583--22599. Association for Computational Linguistics.

\bibitem[{Yue et~al.(2024)Yue, Song, Cheng, and Hu}]{yue2024largelanguagemodelsunderstand}
Shisen Yue, Siyuan Song, Xinyuan Cheng, and Hai Hu. 2024.
\newblock \href {https://aclanthology.org/2024.ccl-1.98/} {Do large language models understand conversational implicature- a case study with a {C}hinese sitcom}.
\newblock In \emph{Proceedings of the 23rd Chinese National Conference on Computational Linguistics (Volume 1: Main Conference)}, pages 1270--1285, Taiyuan, China. Chinese Information Processing Society of China.

\bibitem[{Zheng et~al.(2021)Zheng, Qiu, Fan, Zhu, and Zhu}]{zheng-etal-2021-grice}
Zilong Zheng, Shuwen Qiu, Lifeng Fan, Yixin Zhu, and Song-Chun Zhu. 2021.
\newblock \href {https://doi.org/10.18653/v1/2021.findings-acl.182} {{GRICE}: A grammar-based dataset for recovering implicature and conversational r{E}asoning}.
\newblock In \emph{Findings of the Association for Computational Linguistics: ACL-IJCNLP 2021}, pages 2074--2085, Online. Association for Computational Linguistics.

\bibitem[{Zhou et~al.(2025)Zhou, Karidi, Liu, Garneau, Cao, Chen, Li, and Hershcovich}]{zhou-etal-2025-mapo}
Li~Zhou, Taelin Karidi, Wanlong Liu, Nicolas Garneau, Yong Cao, Wenyu Chen, Haizhou Li, and Daniel Hershcovich. 2025.
\newblock \href {https://doi.org/10.18653/v1/2025.naacl-long.496} {Does mapo tofu contain coffee? probing {LLM}s for food-related cultural knowledge}.
\newblock In \emph{Proceedings of the 2025 Conference of the Nations of the Americas Chapter of the Association for Computational Linguistics: Human Language Technologies (Volume 1: Long Papers)}, pages 9840--9867, Albuquerque, New Mexico. Association for Computational Linguistics.

\end{thebibliography}

\begin{thebibliography}{2}
\expandafter\ifx\csname natexlab\endcsname\relax\def\natexlab#1{#1}\fi

\bibitem[{Floyd et~al.(2023)Floyd, Gibson, Fedorenko, and Poliak}]{floyd_pragmega_resource}
Floyd, Sammy and Gibson, Edward and Fedorenko, Evelina and Poliak, Moshe. 2023.
\newblock \emph{Pragmega}.
\newblock {OSF} repository, Center for Open Science.
\newblock PID \href{https://osf.io/dpge6/}{https://osf.io/dpge6/}.

\bibitem[{Sravanthi et~al.(2024)Sravanthi, Doshi, Tankala, Murthy, Dabre, and Bhattacharyya}]{PUBdataset}
Sravanthi, Settaluri and Doshi, Meet and Tankala, Pavan and Murthy, Rudra and Dabre, Raj and Bhattacharyya, Pushpak. 2024.
\newblock \emph{Pragmatics Understanding Benchmark (PUB)}.
\newblock Hugging Face Hub.
\newblock PID \href{https://huggingface.co/datasets/cfilt/PUB}{https://huggingface.co/datasets/cfilt/PUB}.

\end{thebibliography}

\section{Language Resource References}\label{lr:ref}
\bibliographystylelanguageresource{lrec2026-natbib}
\bibliographylanguageresource{languageresource}

\clearpage
\section*{Appendix A: PragMega Prompts}\label{app:pragmega_prompts}
 \subsection{Irony}
 \paragraph{English}\hfill
 \\
 
 \fbox{{\parbox{0.95\linewidth}{
Task: You will read short stories that describe everyday situations. Each story will be followed by a multiple-choice question. Read each story and choose the best answer. Your task is to decide what the character in the story is trying to convey. The answer options are 1, 2, 3, or 4.

Scenario: \textit{[Example]}

Options:

\textit{[Hypotheses]}

Answer:}}}

\paragraph{Slovene}\hfill
 \\
 
 \fbox{{\parbox{0.95\linewidth}{Naloga: Prebral boš kratko zgodbo, ki opisuje vsakdanjo situacijo. Zgodbi bo sledilo vprašanje in več možnih odgovorov. Preberi zgodbo in izberi najboljši odgovor. Tvoja naloga je, da ugotoviš, kaj je oseba v zgodbi želela sporočiti. Možni odgovori so 1, 2, 3 ali 4.
 
Zgodba: \textit{[Example]}

Možni odgovori:

\textit{[Hypotheses]}

Odgovor:
}}}

\subsection{Metaphor}
 \paragraph{English}\hfill
 \\
 
 \fbox{{\parbox{0.95\linewidth}{Task: You will read short stories that describe everyday situations. Each story will be followed by a multiple-choice question. Read each story and choose the best answer to each question. The answer options are 1, 2, 3, 4, or 5. 
 
Scenario: \textit{[Example]}

Options:

\textit{[Hypotheses]}

Answer:}}}

\paragraph{Slovene}\hfill
 \\

 \fbox{{\parbox{0.95\linewidth}{Naloga: Prebral boš kratko zgodbo, ki opisuje vsakdanjo situacijo. Zgodbi bo sledilo vprašanje in več možnih odgovorov. Preberi zgodbo in izberi najboljši odgovor na vprašanje. Možni odgovori so 1, 2, 3, 4 ali 5.
 
Zgodba: \textit{[Example]} 

Možni odgovori: 

\textit{[Hypotheses]}

Odgovor:}}}

\subsection{Humour}
\paragraph{English}\hfill
 \\

 \fbox{{\parbox{0.95\linewidth}{Task: You will read jokes that are missing their punch lines. A punch line is a funny line that finishes the joke. Each joke will be followed by five possible endings. Please choose the ending that makes the joke funny. The answer options are 1, 2, 3, 4, or 5. 
 
Joke: \textit{[Example]}

Punchlines:

\textit{[Hypotheses]}

Answer:}}}

\paragraph{Slovene}\hfill
 \\
 
 \fbox{{\parbox{0.95\linewidth}{
Naloga: Prebral boš šalo, ki ji manjka zaključek oziroma vrhunec ("punchline"). V tem kontekstu je vrhunec duhovit stavek, ki zaključi šalo. Vsaki šali sledi pet možnih zaključkov. Izberi tisti zaključek, ki kot vrhunec ustvari šalo. Možni odgovori so 1, 2, 3, 4 ali 5.

Šala: \textit{[Example]}

Zaključki:

\textit{[Hypotheses]}

Odgovor:}}}

\end{document}